\documentclass{llncs}

\usepackage[utf8]{inputenc}
\usepackage{amsmath}
\usepackage{graphicx}
\usepackage{amsfonts}
\usepackage{amssymb}
\usepackage{url}

\usepackage{cite}
\usepackage{mathrsfs}  
\usepackage{amssymb,amsfonts,amsmath}
\usepackage{multicol,multicap,graphicx,epstopdf}
\usepackage{graphicx}
\usepackage{subcaption}
\usepackage{wrapfig}
\usepackage{picinpar}
\usepackage{cutwin}
\usepackage{algorithm}  
\usepackage{algorithmic}
\usepackage{verbatim}
\usepackage{url}
\usepackage{hyperref}
\usepackage{epsfig}

\usepackage{textcomp}
\usepackage{xcolor}
\usepackage{stmaryrd}
\usepackage{url}
\DeclareMathOperator*{\argmin}{arg\,min}

\begin{document}

{\let\thefootnote\relax\footnotetext{Copyright \textcopyright\ 2020 for this paper by its authors. Use permitted under Creative Commons License Attribution 4.0 International (CC BY 4.0). CLEF 2020, 22-25 September 2020, Thessaloniki, Greece.}}

\title{Adversarial Consistent Learning on Partial Domain Adaptation of PlantCLEF 2020 Challenge}

\author{Youshan Zhang  \ \and   \ Brian D. Davison}

\institute{Lehigh University, Computer Science and Engineering, Bethlehem, PA, USA\\
\email{\{yoz217,bdd3\}@lehigh.edu}}

\maketitle

\begin{abstract}
Domain adaptation is one of the most crucial techniques to mitigate the domain shift problem, which exists when transferring knowledge from an abundant labeled sourced domain to a target domain with few or no labels. Partial domain adaptation addresses the scenario when target categories are only a subset of source categories.  In this paper, to enable the efficient representation of cross-domain plant images,  we first extract deep features from pre-trained models and then develop adversarial consistent learning ($ACL$) in a unified deep architecture for partial domain adaptation. It consists of source domain classification loss, adversarial learning loss, and feature consistency loss. Adversarial learning loss can maintain domain-invariant features between the source and target domains. Moreover, feature consistency loss can preserve the fine-grained feature transition between two domains. We also find the shared categories of two domains via down-weighting the irrelevant categories in the source domain. Experimental results demonstrate that training features from NASNetLarge model with proposed $ACL$ architecture yields promising results on the PlantCLEF 2020 Challenge.
\end{abstract}

\section{Introduction}
 Automated plant identification is important in recognizing plant species. The availability of massive labeled training data is a prerequisite of machine learning models. Unfortunately,  such a  requirement cannot be met in the plant identification problem since we have sparse labels for real-world plant images. Therefore, we propose to transfer knowledge from an existing auxiliary labeled herbarium domain to the field photo domain with limited or no labels. However, due to the phenomenon of data bias or domain shift \cite{pan2010survey}, classification models do not generalize well from an existing herbarium domain to a novel field photo domain. 

Domain adaptation (DA) has been proposed to leverage knowledge from an abundant labeled source domain to learn an effective predictor for the target domain with few or no labels, while mitigating the domain shift problem~\cite{zhang2019modified,zhang2019transductive,zhang2020impact,zhang2020domain}. In this paper, we focus on unsupervised domain adaptation (UDA), where the target domain has no labels. Since we have fewer classes in the field photo domain, and the classes of the field photo domain is a subset of the classes of the source herbarium domain, we investigate partial domain adaptation (PDA) for the PlantCLEF 2020 Challenge. 

Recently, deep neural network methods have been widely used in the domain adaptation problem. Notably, adversarial learning shows its power in embedding in deep neural networks to learn feature representations to minimize the discrepancy between the source and target domains~\cite{tzeng2017adversarial,liu2019transferable}. Inspired by the generative adversarial network (GAN)~\cite{goodfellow2014generative}, adversarial learning also contains a feature extractor and a domain discriminator. The domain discriminator can distinguish the source domain from the target domain, while the feature extractor can learn domain-invariant representations to fool the domain discriminator \cite{long2018conditional,zhang2019domain,liu2019transferable}. The target domain risk (the error of the target domain)  is expected to be minimized via minimax optimization. Cao et al.\ presented adversarial learning for PDA, which alleviates negative transfer by reducing the outlier of source classes for training the source classifier and domain labels, while positive transfer is improved via matching the feature distributions in the shared label space~\cite{cao2018partial}. Similarly, the example transfer network is proposed to jointly learn domain-invariant representations and a progressive weighting method to examine the transferability of source examples. The model can improve positive transfer by relevant examples and mitigate negative transfer by identifying irrelevant examples~\cite{chen2019domain}.

Although many methods are proposed for partial domain adaptation, they still suffer from two challenges: (1) the models are evaluated on small datasets, while it has lower transferability to the large-scale dataset, and (2) the feature consistency of two domains is inappropriately ignored. 

To address the aforementioned challenges, we aggregate three different loss functions in one framework: source domain classification loss, adversarial learning loss, and feature consistency loss to reduce the discrepancy of the two domains. Moreover, our model is evaluated on a large-scale plant identification dataset to improve the estimate of the generalization ability of our model. 

Our contributions are three-fold:
\begin{enumerate}
    \item We propose a novel adversarial consistent learning network ($ACL$) for PDA, to adversarially minimize the domain discrepancy of the source and target domains and maintain domain-invariant features;
    \item The proposed adversarial learning loss and feature consistency loss can 
    distinguish the target domain from the source domain, and  preserve the fine-grained feature transition between the two domains;
    \item We 
    impose shared category selection to filter out the irrelevant categories in the source domain. By down-weighting the irrelevant categories in the source domain, we can reduce negative transfer from the source domain to the target domain. 
\end{enumerate}
Experimental results show that $ACL$ achieves higher classification accuracy than several baseline methods and yields promising results on the PlantCLEF 2020 Challenge.

\section{Dataset}
\begin{figure}[t]
\centering
\includegraphics[scale=0.23]{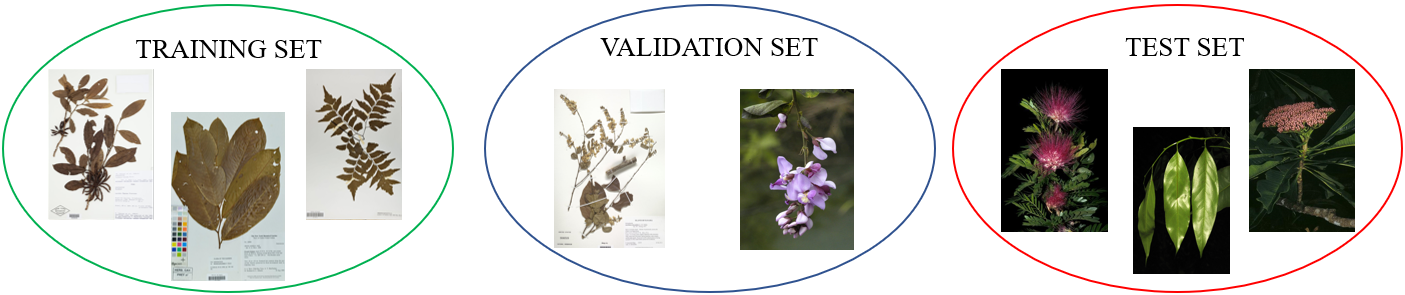}
\caption{Examples images of the PlantCLEF 2020 dataset.  The large discrepancy between training and test data cause the difficulty in the PDA.  }
\label{fig:plant_eax}
\vspace{-0.3cm}
\end{figure}

\textbf{PlantCLEF 2020} is a large-scale dataset of the PlantCLEF 2020 task~\cite{plantclef2020}, organized in the context of the LifeCLEF 2020 challenge~\cite{lifeclef2020}. Fig.~\ref{fig:plant_eax} shows some challenging images in this dataset. The herbarium domain contains 320,750 images in 997 species, and the number of images in different species are unbalanced. This dataset consists of herbarium sheets whereas the test set will be composed of field pictures. The validation set consists of two domains herbarium\_photo\_associations and photos. Herbarium\_photo\_associations domain includes 1,816 images from 244 species. This domain contains both herbarium sheets and field pictures for a subset of species, which enables learning a mapping between the herbarium sheets domain and the field pictures domain. Another photo domain has 4,482 images from 375 species and images are from plant pictures in the field, which is similar to the test dataset. The test dataset contains 3,186 unlabeled images. Due to the significant difference between herbarium and real photos, it is extremely difficult to identify the correct class.
\begin{table}[b]
\footnotesize
\begin{center}
\caption{Statistics on PlantCLEF 2020 dataset}
\setlength{\tabcolsep}{+3.2mm}{
\begin{tabular}{rccccccccccccc}
\hline \label{tab:data}
Domain &  Number of Samples & Number of Classes \\
\hline
Herbarium (H) &  320,750 & 997 \\
Herbarium\_photo\_associations (A)& 1,816 &244\\
Photo (P)& 4,482  & 375 \\
Test (T) & 3,186   &- \\
 \hline
\end{tabular}}
\end{center}
\end{table}

We exclude the classId of ``108335" in the photo domain since the major classes are from the herbarium domain. In addition, herbarium domain does not contain the ``108335" category. Therefore, eight images are excluded in the photo domain. The statistics of the PlantCLEF 2020 dataset are listed in Tab.~\ref{tab:data}.

\begin{figure}[t]
\centering
\includegraphics[scale=0.25]{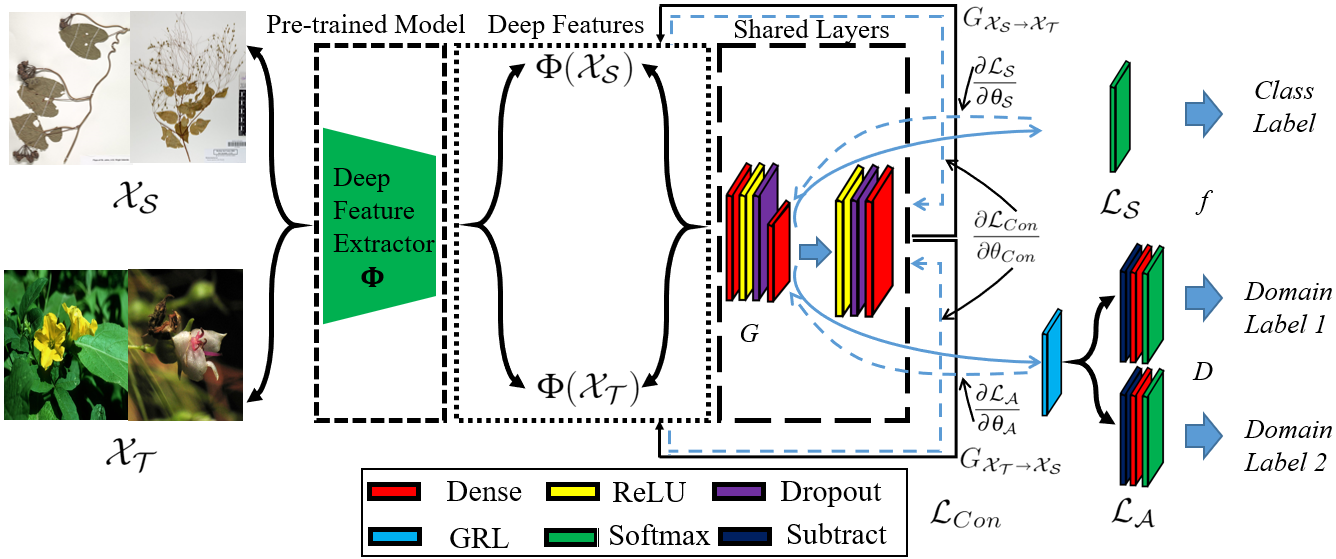}
\caption{The architecture of our proposed $ACL$ model. We first extract deep features from a pre-trained model for both source and target domains via $\Phi$. The shared layers are jointly trained with source and target features. Also, the parameters in shared layers are updated by the backward gradients ($\frac{\partial \mathcal{L_S}}{\partial \theta_\mathcal{S}}$, $\frac{\partial \mathcal{L_A}}{\partial \theta_\mathcal{A}}$ and $\frac{\partial \mathcal{L}_{Con}}{\partial \theta_{Con}}$) from class label classifier, domain label predictor  and feature consistency regressor. The $ACL$ model consists of three different loss functions (source classification loss $\mathcal{L_S}$, adversarial domain loss $\mathcal{L_A}$, and  feature consistency loss $\mathcal{L}_{Con}$). The feature extractor $G$ in the shared layers is used for both classifier $f$ and domain discriminator $D$ (The blue dash lines are the backward gradients, and GRL stands for gradient reversal layer). Layers visualization of architecture is shown in Fig.~\ref{fig:layer}.  }
\label{fig:plant}
\end{figure}

\section{Methods}

\subsection{Motivation}

Previous partial domain adaptation methods~\cite{cao2018partial,chen2019domain} evaluated their models based on a small dataset (e.g., Office 31), while their models have lower generalizability to large-scale datasets. In addition, feature consistency of both source and target domains is not well addressed in the PDA. 

In this paper, we present our approach: adversarial consistent learning (ACL) on partial domain adaptation.  It can align the feature distribution of the source and target domains in the shared categories and guarantee feature consistency across the two domains. Importantly, ACL identifies irrelevant source
categories via down-weighting class importance automatically. Evaluation on the large-scale PlantCLEF 2020 challenge dataset shows a high  generalizability of our model.

\subsection{Problem and notation}

For unsupervised domain adaptation, given a source domain $\mathcal{D_S} = \{\mathcal{X_S}_i, \mathcal{Y_S}_i \}_{i=1}^{\mathcal{N_S}}$ of $\mathcal{N_S}$ labeled samples across the set of categories $\mathcal{C_S}$ and a target domain $\mathcal{D_T} = \{\mathcal{X_T}_j\}_{j=1}^{\mathcal{N_T}}$ of $\mathcal{N_T}$ samples without any labels ($\mathcal{Y}_\mathcal{T}$ is unknown) across the set of categories $\mathcal{C_T}$. For partial domain adaptation, the number of categories in $\mathcal{C_T}$ is less than  the number of categories in $\mathcal{C_S}$, and $\mathcal{C_T} \subsetneqq \mathcal{C_S}$. The samples $\mathcal{X_S}$ and $\mathcal{X_T}$  obey the marginal distributions of $P_{\mathcal{S}}$ and $P_{\mathcal{T}}$. The conditional distributions of two domains are denoted as $Q_{\mathcal{S}}$ and  $Q_{\mathcal{T}}$. Due to the discrepancy of two domains, the distributions are assumed to be different, i.e.,  $P_{\mathcal{S}} \neq P_{\mathcal{T}}$ and $Q_{\mathcal{S}} \neq Q_{\mathcal{T}}$.  Our ultimate goal is to learn a classifier $f$ under a feature extractor $G$, which selects shared categories between two domains, and ensures lower generalization error in the target domain.

\subsection{Deep features extraction}
To circumvent the large computation resource requirement of training large-scale PlantCLEF 2020 challenge datasets, we instead focus on deep features from pre-trained models. Based on Zhang and Davison \cite{zhang2020impact}, the deep features are extracted from the last fully connected layer of the pretrained model via $\Phi$. One represented feature vector has the size of $1\times 1000$ and corresponds to one plant image. Therefore,  the source domain and the target domain can be represented by $\Phi(\mathcal{{X_{S}}}) \in \mathbb{R}^{\mathcal{N}_\mathcal{S} \times 1000}$ and $\Phi(\mathcal{{X_{T}}}) \in \mathbb{R}^{\mathcal{N}_\mathcal{T} \times 1000}$, respectively. 

\subsection{Source classifier}
The task in the source domain is trained using the typical cross-entropy loss in following equation:
\begin{equation}\label{eq:lc}
   \mathcal{L_S} (f(G(\Phi(\mathcal{X_{S}}))),\mathcal{Y_{S}}) = - \frac{1}{\mathcal{N_S}}\sum_{i=1}^{\mathcal{N_S}} \sum_{c=1}^{\mathcal{C_S}} \mathcal{Y_{S}}_{ic} \text{log}  (f(G(\Phi(\mathcal{X_{S}}_i)))), 
\end{equation}
where $\mathcal{Y_{S}}_{ic} \in [0, 1]^{\mathcal{C_S}}$ is the probability of each class of ground truth for the $i$th element of S, $f$ is the classifier in Fig.~\ref{fig:plant}, and $f(G(\Phi(\mathcal{X_{S}}_i)))$ is the predicted probability.

\subsection{Adversarial domain loss}
In general adversarial learning, the system learns a mapping from the source domain to the target domain. Given the feature representation of feature extractor $G$, we can learn a discriminator $D$, which can distinguish the two domains using the following loss function:
\begin{equation}
\begin{aligned}\label{eq:G1}
    \mathcal{L_A}(G_{\mathcal{X_S}\shortrightarrow \mathcal{X_T}}, G(\Phi(\mathcal{X_S})), G(\Phi(\mathcal{X_T})))  = & - \frac{1}{ \mathcal{N_S}} \sum_{i=1}^{ \mathcal{N_S}} \text{log} (1-D(G(\Phi(\mathcal{X_{S}}_i))))  \\ &  - \frac{1}{ \mathcal{N_T}} \sum_{j=1}^{\mathcal{N_T}} \text{log} (D(G(\Phi(\mathcal{X_T}_j)))). 
\end{aligned}
\end{equation}
However, Eq.~\ref{eq:G1} only guarantees source domain data will be close to the target data ($G_{\mathcal{X_S}\shortrightarrow \mathcal{X_T}}$), and it does not ensure that the target data will be close to the source data. We hence introduce another mapping from the target domain to the source domain $G_{\mathcal{X_T}\shortrightarrow \mathcal{X_S}}$ in Eq.~\ref{eq:G2} and train it with the same adversarial loss as in  $G_{\mathcal{X_S}\shortrightarrow \mathcal{X_T}}$ as shown in Eq.~\ref{eq:G1}. 
\begin{equation}
\begin{aligned}\label{eq:G2}
    \mathcal{L_A}(G_{\mathcal{X_T}\shortrightarrow \mathcal{X_S}}, G(\Phi(\mathcal{X_S})), G(\Phi(\mathcal{X_T})))
\end{aligned}
\end{equation}
For $G_{\mathcal{X_S}\shortrightarrow \mathcal{X_T}}$, the source domain has the label of $0$ and the target domain has the label of $1$, which is corresponding to \textit{Domain Label 1} in Fig.~\ref{fig:plant}. Meanwhile, for $G_{\mathcal{X_T}\shortrightarrow \mathcal{X_S}}$, $1$ is the new label for the source domain and and $0$ is the new label for target domain, which is corresponding to \textit{Domain Label 2} in Fig.~\ref{fig:plant}. Therefore, we define the adversarial learning loss as:
\begin{equation}
\begin{aligned}\label{eq:A}
   \mathcal{L_{A}} (G(\Phi(\mathcal{X_S})), G(\Phi(\mathcal{X_T}))) & =     \mathcal{L_A}(G_{\mathcal{X_S}\shortrightarrow \mathcal{X_T}}, G(\Phi(\mathcal{X_S})), G(\Phi(\mathcal{X_T}))) \\&+  \mathcal{L_A}(G_{\mathcal{X_T}\shortrightarrow \mathcal{X_S}}, G(\Phi(\mathcal{X_S})), G(\Phi(\mathcal{X_T}))).
\end{aligned}
\end{equation}

\subsection{Feature consistency loss}

To encourage the source domain and target domain information to be preserved during adversarial learning, we propose a feature consistency loss in our model. Details of the feature reconstruction layers are shown in Fig.~\ref{fig:plant}; the reconstructed layers are right behind the feature extractor $G$ in the shared layers, and they aim to reconstruct the extracted features and maintain the invariant features during the conversion process. The feature consistency loss is defined as:

\begin{equation}
\begin{aligned}\label{eq:all1}
    \mathcal{L}_{Con} & (G_{\mathcal{X_S}\shortrightarrow \mathcal{X_T}}, G_{\mathcal{X_T}\shortrightarrow \mathcal{X_S}}, G(\Phi(\mathcal{X_S})), G(\Phi(\mathcal{X_T}))) \\ &= \mathbb{E}_{x_{s} \sim G(\Phi(\mathcal{X_S})) }  [\ell(G_{\mathcal{X_T}\shortrightarrow \mathcal{X_S}} (G_{\mathcal{X_S}\shortrightarrow \mathcal{X_T}} (x_s)) - x_s )] \\ &
    + \mathbb{E}_{x_{t} \sim G(\Phi(\mathcal{X_T})) }  [\ell(G_{\mathcal{X_S}\shortrightarrow \mathcal{X_T}} (G_{\mathcal{X_T}\shortrightarrow \mathcal{X_S}} (x_{t})) - x_{t} )],
\end{aligned}
\end{equation}
where $\ell$ is the mean squared error loss function, which calculates the difference between true features and the reconstructed features. 

\subsection{Shared categories selection}
In PDA, the set of target domain labels is a subset of the source domain labels, i.e., $\mathcal{C_T}  \subsetneqq  \mathcal{C_S}$. In the PlantCLEF challenge, the size of irrelevant label set ($\mathcal{C_S}-\mathcal{C_T}$) is far larger than the size of $\mathcal{C_T}$ ($|\mathcal{C_S}-\mathcal{C_T}| >> |\mathcal{C_T}|$). If we use all elements of the source domain distribution to match the target domain distribution, it will cause  negative transfer since the target domain will also be forced to match the irrelevant labels ($\mathcal{C_S}-\mathcal{C_T}$). Therefore, it is important to identify the shared categories between source and target domains. 

To address the aforementioned challenge, we re-weight the source domain label set via reducing the irrelevant label set. During the training, we can get the predicted probability of the target domain: $\hat{\mathcal{Y_{T}}_{j}}= f(G(\Phi(\mathcal{X_{T}}_j)))$, which gives a probability of each source label in $\mathcal{C_S}$. As we know, the set of irrelevant source labels and
target label set are disjoint, and  the target data are significantly dissimilar to the source data in the irrelevant label set. Therefore, the probability of irrelevant categories should be sufficiently small and can be ignored. We then defined the weight vector as:
\begin{equation}
\begin{aligned}\label{eq:w}
    \mathcal{W} = \frac{1}{\mathcal{N_T}} \sum_{j=1}^{\mathcal{N_T}}  \hat{\mathcal{Y_{T}}_{j}},
\end{aligned}
\end{equation}
where $\mathcal{W}$ is a $|\mathcal{C_S}|$-dimensional weight vector. The irrelevant categories ($\mathcal{C_S}-\mathcal{C_T}$) will have a much smaller weight than the shared categories. We then assign the weight as $0$ if its element $\mathcal{W}_c$ is less than a sufficiently small number (e.g., $10e-9$). By reducing the weight of irrelevant categories, the shared categories can be emphasized and negative transfer will be mitigated. The weight vector $\mathcal{W}$ is applied in both the source 
classifier and   domain discriminator over the source domain data as shown in the following objective function.

\subsection{Overall objective}
We combine the three aforementioned loss functions to formalize our objective function:
\begin{equation}
\begin{aligned}\label{eq:loss_all}
 & \mathcal{L} (\mathcal{X_S}, \mathcal{X_T}, \mathcal{Y_S}, G_{\mathcal{X_S}\shortrightarrow \mathcal{X_T}}, G_{\mathcal{X_T}\shortrightarrow \mathcal{X_S}} ) \\ & = \mathcal{L_S} (f(\mathcal{W}(G(\Phi(\mathcal{X_{S}})))),\mathcal{Y_{S}})+ \gamma  \mathcal{L_{A}} (\mathcal{W}(G(\Phi(\mathcal{X_S}))), G(\Phi(\mathcal{X_T})))   \\& + \beta \mathcal{L}_{Con} (G_{\mathcal{X_S}\shortrightarrow \mathcal{X_T}}, G_{\mathcal{X_T}\shortrightarrow \mathcal{X_S}}, \mathcal{W}(G(\Phi(\mathcal{X_S}))), G(\Phi(\mathcal{X_T}))),
\end{aligned}
\end{equation}
where $\gamma$ and $\beta$ are tradeoff parameters between different loss functions. Our model ultimately  minimizes the difference during the transition from the source domain to target domain and from the target domain to the source domain. Meanwhile, it maximizes the ability to distinguish the two domains.  


\subsection{Gradients of shared layers}
The shared layers consist of the feature extractor $G$ and the feature reconstruction layers. In $G$, there are two dense layers, a ``Relu" activation layer, and a dropout layer. The numbers of units of the dense layer are 1000 and 997, respectively. The rate of the Dropout layer is 0.5. The feature reconstruction layers  have a ``Relu" activation layer, a dropout layer and a dense layer with the number of units of 1000.  The shared layers are jointly optimized by both the source classification loss, adversarial domain loss and feature consistency loss. 

Let $\mathcal{F_E} ( \cdot,\theta_\mathcal{E})$ be the output of the shared encoder with parameters of $\theta_\mathcal{E}$. In addition, let $\mathcal{F_S} ( \cdot,\theta_\mathcal{S})$ be the output of class label classifier with parameters of $\theta_\mathcal{S}$, $\mathcal{F_A} ( \cdot,\theta_\mathcal{A})$ be the output of domain label predictor with parameters of $\theta_\mathcal{A}$, and $\mathcal{F}_{Con} ( \cdot,\theta_{Con})$ be the  output of feature consistency regressor with parameters of $\theta_{Con}$. Therefore, the shared layers are optimized by these three gradients. The parameters in the shared layers are updated in the following equation:

\begin{equation}
\begin{aligned}\label{eq:share}
\theta_\mathcal{S} \shortleftarrow \theta_\mathcal{S} - \eta\frac{\partial \mathcal{L_S}}{\partial \theta_\mathcal{S} }, & \ \theta_\mathcal{A}  \shortleftarrow \theta_\mathcal{A}- \eta\tau\theta_\mathcal{A} \frac{\partial \mathcal{L_A}}{\partial \theta_\mathcal{A} }, \   \theta_{Con} \shortleftarrow  \theta_{Con} - \eta  \frac{\partial \mathcal{L}_{Con}}{\partial \theta_{Con}}
\\ & \ \theta_\mathcal{E} \shortleftarrow \theta_\mathcal{E} - \eta (\frac{\partial \mathcal{L_S}}{\partial \theta_\mathcal{S}} + \tau\theta_\mathcal{A} \frac{\partial \mathcal{L_A}}{\partial \theta_\mathcal{A} } + \frac{\partial \mathcal{L}_{Con}}{\partial \theta_{Con}}),  
\end{aligned}
\end{equation}
where $\eta$ is the learning rate and $\tau$ is the adaptation factor from gradient reversal layer (GRL) in~\cite{ghifary2014domain}.

\subsection{Theoretical Analysis}

We now formalize the error bound of our model. $ACL$ model is trained with both the labeled source domain and the unlabeled target domain. The error bound of the source domain and the target domain ($\epsilon (h)$) in our model is then formally written as:
\begin{equation}
\begin{aligned}\label{eq:analysis}
\epsilon (h) & = \epsilon_{\mathcal{X_S}} (h, \mathcal{Y_S}) + \epsilon_{\mathcal{X_T}} (h, \Hat{\mathcal{Y_T}}),
\end{aligned}
\end{equation}
where $\Hat{\mathcal{Y_T}}$ is the predicted label of target domain. The term $\epsilon_{\mathcal{X_S}} (h, \mathcal{Y_S}) = \mathbb{E}_{x \sim \mathcal{X_S}} [|h (x) - \mathcal{Y_S}|]$ and $\epsilon_{\mathcal{X_T}} (h, \Hat{\mathcal{Y_T}}) = \mathbb{E}_{x \sim \mathcal{X_T}} [|h (x) - \Hat{\mathcal{Y_T}}|]$ denote the expected risk over the source domain and the target domain with respect to the ground truth labels and  predicted labels, respectively (where $|\cdot|$ is the L1 norm).  
 
During the training, we expect the error $\epsilon_{\mathcal{X_T}} (h, \Hat{\mathcal{Y_T}}))$ to be close to $\epsilon_{\mathcal{X_T}} (h, \mathcal{Y_T})$, which evaluates the classifier $f$ with true target domain labels.  The smaller the difference between these two errors, the better the model performs and more discrepancies of the two domains are reduced.  Existing domain adaptation theory shows that the risk in the target domain can be minimized by bounding the source risk and discrepancy between source and target domains~\cite{ben2010theory}). Therefore, the generalization error bound of our model is shown in the following Lemma.

\textbf{Lemma 1}  Let $h$ be a hypothesis in a class $H$. Then

\begin{equation}
\begin{aligned}\label{eq:ana}
\epsilon (h) & = \epsilon_{\mathcal{X_S}} (h, \mathcal{Y_S}) + \epsilon_{\mathcal{X_T}} (h, \Hat{\mathcal{Y_T}}) \\ & 
\leq 2\epsilon_{\mathcal{X_S}} (h) + d_{\mathcal{H}} (\mathcal{D_S}, \mathcal{D_T}) + C,
\end{aligned}
\end{equation}
where $d_{\mathcal{H}} (\mathcal{D_S}, \mathcal{D_T}) = 2 \sup _{h, h' \in H} |\epsilon_{\mathcal{X_S}} (h, h') - \epsilon_{\mathcal{X_T}} (h, h')|$ is the $\mathcal{H}$-divergence of training and test data in the hypothesis space $H.$  $C = \epsilon_{\mathcal{X_S}} (h^{*}, \mathcal{Y_S}) + \epsilon_{\mathcal{X_T}} (h^{*}, \mathcal{Y_T})$ is the adaptability to quantify the error in ideal hypothesis $h^{*}$ space of training and test data, which should be small and is the optimal hypothesis via minimizing the joint error in Eq.~\ref{eq:h}.

\begin{equation}\label{eq:h}
    h^{*} = \argmin \epsilon_{\mathcal{X_S}} (h, \mathcal{Y_S}) + \epsilon_{\mathcal{X_T}} (h, \mathcal{Y_T})
\end{equation}

In Lemma 1, the generalization boundary of our model consists of three terms: training data error, data discrepancy $d_{\mathcal{H}} (\mathcal{D_S}, \mathcal{D_T})$, which is estimated by the disagreement of hypothesis in the space $H$, and the adaptability $C$ of the ideal joint hypothesis. In $ACL$ model, the first term is measured by Eq.~\ref{eq:lc}. The domain discrepancy is assessed by adversarial learning loss and feature consistency loss.  Furthermore,  $ACL$ finds the ideal hypothesis and reduces the training error in each iteration.  Hence, our model can find a minimal boundary for two domains. In other words,  $ACL$ can implicitly minimize the target domain risk, domain discrepancy, and the adaptability of true hypothesis $h$ in terms of the hypothesis space $H.$ 

 \begin{figure}[t]
\centering
\includegraphics[width=1.0\columnwidth]{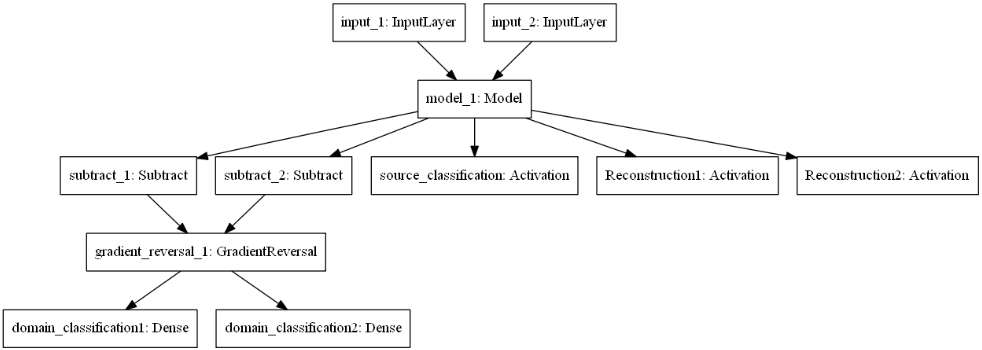}
\caption{Layers visualization of our proposed $ACL$ model.  Two input layers are from source domain data and target domain data, respectively. The intermediate model is the shared layers in Fig.~\ref{fig:plant}. The source classification layer refers to the classifier $f$, and two reconstruction layers guarantee the feature consistency of two domains. Two subtract layers are used for the domain discriminator. In addition, the gradient reversal layer is used for backpropagation.  }
\label{fig:layer}
\end{figure}

\section{Experiments}
\subsection{Implementation details}
As aforementioned, the deep features are extracted from the last fully connected layer~\cite{zhang2018automated,zhang2020impact}. One represented feature vector has the size of $1\times 1000$ and is corresponding to one plant image. Therefore, the feature representation of domain herbarium (H) has the size of $320,750 \times 1000$, domain herbarium\_photo\_associations (A) has the size of $1,816 \times 1000$, domain photo (P) has the size of  $4,482  \times 1000$, and domain test (T) has the size of $3,186 \times 1000$. In the experiment, our task is to reduce the error in the target domain (real-world plant images), i.e., photo domain or test domain. Our tasks will focus more on the evaluation of the domain P and domain T. Since the herbarium\_photo\_associations (A) is important to bridge the map between two domains, we hence include the domain A in the training procedure to form a new source domain, which consists of domain herbarium (H)  and domain A. Domain H + A has the size of $322,566 \times 1000$. We then train the model based on these extracted feature vectors. In Tab.~\ref{tab:P1},  H $\shortrightarrow$ P represents learning knowledge from domain H, which is applied to domain P. 

The parameters of $ACL$ are first tuned based on the performance of the domain P, while the model is trained with H + A domain. We then apply these parameters to domain T and submit it to the challenge for the evaluation. Our implementation is based on Keras. The parameters settings are  $\beta=\gamma = 0.5$, $\tau =0.31$, learning rate: $\eta= 0.0001$, batch size = 128, the number of iterations is 1000 and the optimizer is Adam. The details of the layers are shown in Fig.~\ref{fig:layer}. 

We also compare our results with two domain adaptation methods:  DANN~\cite{ghifary2014domain} and ADDA~\cite{tzeng2017adversarial}. In addition, we extracted features from four well-trained models (ResNet50~\cite{he2016deep}, InceptionV3~\cite{szegedy2016rethinking}, Inception-Resnet-V2~\cite{szegedy2017inception}, NASNetLarge~\cite{zoph2018learning}), which is trained based on large-scale ImageNet datasets. We then feed these different extracted features into the shared layers and optimize the objective function in Eq.~\ref{eq:loss_all}.

\subsection{Results}
\begin{table}[t]
\footnotesize
\begin{center}
\caption{Accuracy (\%) on PlantCLEF 2020 dataset  for photo domain}
\setlength{\tabcolsep}{+2.7mm}{
\begin{tabular}{rccccccccccccc}
\hline \label{tab:P1}
Task & A $\shortrightarrow$ P & H $\shortrightarrow$ P & H+A $\shortrightarrow$ P  \\
\hline
DANN~\cite{ghifary2014domain} & 1.07  & 1.85 &  2.01 \\
ADDA~\cite{tzeng2017adversarial} &2.95  & 3.05 &  3.43\\
\hline
\hline
ResNet50-$ACL$ &2.96 & 4.83& 6.97\\
InceptionV3-$ACL$ &3.02 & 5.93& 7.95\\
Inception-Resnet-V2-$ACL$ &3.73 & 7.07 & 8.43 \\
\textbf{NASNetLarge-$ACL$ $-\mathcal{W}$} & 3.84 & 7.92 & 8.18 \\
\textbf{NASNetLarge-$ACL$} & \textbf{5.98} & \textbf{8.64} & \textbf{9.67} \\
 \hline
\end{tabular}}
\end{center}
\vspace{-0.6cm}
\end{table}
The performance of the photo domain is shown in Tab.~\ref{tab:P1}. We report the accuracy of the whole photo domain ($Acc = \sum_{j=1}^{\mathcal{N_T}} (\Hat{\mathcal{Y_T}_j} == \mathcal{Y_T}_j) / \mathcal{N_T} \times 100)$, where $\Hat{\mathcal{Y_T}}$ is the predicted label for the target domain. We can observe that the extracted features from NASNetLarge with our $ACL$ architecture achieves the highest performance across all three tasks. We observe that two domain adaptation methods have relatively lower performance in all three tasks. One reason is that these two methods have weak feature extractors, and they do not exclude the irrelevant categories in the source domain, which might cause the negative transfer. Moreover, with the increasing of the ImageNet model, we can extract better features from plant images, which lead to the high performance of the NASNetLarge-$ACL$ model. In addition, we conduct an ablation study in which we train the best NASNetLarge-$ACL$ model without the shared categories selection (NASNetLarge-$ACL$ $-\mathcal{W}$). The results from all three tasks are lower than NASNetLarge-$ACL$ model, which indicates the shared categories selection is useful in our model.  These experiments demonstrate the efficiency of the $ACL$ model in finding the invariant-features of two domains. 

In the final stage of the PlantCLEF 2020 Challenge, our solutions are evaluated by the organizers using the test domain data. As shown in Tab.~\ref{tab:P2}, our method achieved mean reciprocal rank (MRR) of 0.032 in the whole test domain, and MRR of 0.016 in the subset of the test domain, and our method places 4th in the contest.  

\begin{table}[t]
\footnotesize
\begin{center}
\caption{MRR on PlantCLEF 2020 dataset  for test domain~\cite{plantclef2020}}
\setlength{\tabcolsep}{+4.2mm}{
\begin{tabular}{rccccccccccccc}
\hline \label{tab:P2}
Team & Full test set & Sub-set of the test set   \\
\hline
ITCR PlantNet & 0.180 &	0.052 \\
Neuon AI &0.121	&0.107 \\
UWB &0.039 &	0.007\\
\hline
\hline
\textbf{LU(ours)} &0.032 & 0.016\\
\hline
\hline
Domain &0.031 &	0.015 \\
To Be & 0.028 &	0.016  \\
SSN &0.008 &	0.003\\
 \hline
\end{tabular}}
\end{center}
\vspace{-0.7cm}
\end{table}
\vspace{-0.3cm}
\section{Discussion}
There are two compelling advantages of the $ACL$ model. First, we consider the adversarial consistent learning paradigm, which maintains the domain-invariant features from the source domain to the target domain and vice versa. Secondly, we reduce the weight of irrelevant categories in the source domain, which eliminates the negative transfer during the training.  Although the performance of our model is better than several baseline methods, the highest accuracy of the photo domain is less than 10\%, which illustrates that the transfer learning ability in the real world image is lower. One underlying reason is that  PlantCLEF 2020 Challenge has difficult datasets---that there are significant differences between herbarium domain and photo domain, as shown in Fig.~\ref{fig:plant_eax}. Another reason is caused by the weakness of our model since we only train deep features instead of raw images to reduce the computational requirements; some features might be ignored during the training. The performance of the $ACL$ model could be improved if we train the architecture with raw images.

\section{Conclusion}
In this paper, we propose an adversarial consistent learning network on partial domain adaptation termed ($ACL$) to overcome limitations in finding proper shared categories and guaranteeing the feature consistency of two domains. Our model is optimized via minimizing a three-component loss function. As each component of our $ACL$ model, explicit domain-invariant features are maintained through such a cross-domain training scheme.  Experimental results demonstrate our proposed $ACL$ model yields promising results on the PlantCLEF 2020 Challenge. 

%
%
%
\scriptsize
\bibliographystyle{splncs04}
\bibliography{mybibliography}

\begin{thebibliography}{10}
\providecommand{\url}[1]{\texttt{#1}}
\providecommand{\urlprefix}{URL }
\providecommand{\doi}[1]{https://doi.org/#1}

\bibitem{ben2010theory}
Ben-David, S., Blitzer, J., Crammer, K., Kulesza, A., Pereira, F., Vaughan,
  J.W.: A theory of learning from different domains. Machine Learning
  \textbf{79}(1-2),  151--175 (2010)

\bibitem{cao2018partial}
Cao, Z., Ma, L., Long, M., Wang, J.: Partial adversarial domain adaptation. In:
  Proceedings of the European Conference on Computer Vision (ECCV). pp.
  135--150 (2018)

\bibitem{chen2019domain}
Chen, J., Wu, X., Duan, L., Gao, S.: Domain adversarial reinforcement learning
  for partial domain adaptation. arXiv preprint arXiv:1905.04094  (2019)

\bibitem{ghifary2014domain}
Ghifary, M., Kleijn, W.B., Zhang, M.: Domain adaptive neural networks for
  object recognition. In: Pacific Rim international conference on artificial
  intelligence. pp. 898--904. Springer (2014)

\bibitem{plantclef2020}
Go\"{e}au, H., Bonnet, P., Joly, A.: Overview of lifeclef plant identification
  task 2020. In: CLEF working notes 2020, CLEF: Conference and Labs of the
  Evaluation Forum, Sep. 2020, Thessaloniki, Greece. (2020)

\bibitem{goodfellow2014generative}
Goodfellow, I., Pouget-Abadie, J., Mirza, M., Xu, B., Warde-Farley, D., Ozair,
  S., Courville, A., Bengio, Y.: Generative adversarial nets. In: Advances in
  neural information processing systems. pp. 2672--2680 (2014)

\bibitem{he2016deep}
He, K., Zhang, X., Ren, S., Sun, J.: Deep residual learning for image
  recognition. In: Proceedings of the IEEE Conference on Computer Vision and
  Pattern Recognition (CVPR). pp. 770--778 (2016)

\bibitem{lifeclef2020}
Joly, A., Go\"{e}au, H., Kahl, S., Deneu, B., Servajean, M., Cole, E., Picek,
  L., Ruiz De Casta\~{n}eda, R., {\'e}, Lorieul, T., Botella, C., Glotin, H.,
  Champ, J., Vellinga, W.P., St{\"o}ter, F.R., Dorso, A., Bonnet, P.,
  M\"{u}ller, H.: Overview of lifeclef 2020: a system-oriented evaluation of
  automated species identification and species distribution prediction. In:
  Proceedings of CLEF 2020, CLEF: Conference and Labs of the Evaluation Forum,
  Sep. 2020, Thessaloniki, Greece. (2020)

\bibitem{liu2019transferable}
Liu, H., Long, M., Wang, J., Jordan, M.: Transferable adversarial training: A
  general approach to adapting deep classifiers. In: International Conference
  on Machine Learning. pp. 4013--4022 (2019)

\bibitem{long2018conditional}
Long, M., Cao, Z., Wang, J., Jordan, M.I.: Conditional adversarial domain
  adaptation. In: Advances in Neural Information Processing Systems. pp.
  1647--1657 (2018)

\bibitem{pan2010survey}
Pan, S.J., Yang, Q., et~al.: A survey on transfer learning. IEEE Transactions
  on knowledge and data engineering  \textbf{22}(10),  1345--1359 (2010)

\bibitem{szegedy2017inception}
Szegedy, C., Ioffe, S., Vanhoucke, V., Alemi, A.A.: Inception-v4,
  inception-resnet and the impact of residual connections on learning. In:
  Thirty-First AAAI Conference on Artificial Intelligence (2017)

\bibitem{szegedy2016rethinking}
Szegedy, C., Vanhoucke, V., Ioffe, S., Shlens, J., Wojna, Z.: Rethinking the
  inception architecture for computer vision. In: Proceedings of the IEEE
  conference on computer vision and pattern recognition. pp. 2818--2826 (2016)

\bibitem{tzeng2017adversarial}
Tzeng, E., Hoffman, J., Saenko, K., Darrell, T.: Adversarial discriminative
  domain adaptation. In: Proceedings of the IEEE Conference on Computer Vision
  and Pattern Recognition. pp. 7167--7176 (2017)

\bibitem{zhang2018automated}
Zhang, Y., Allem, J.P., Unger, J.B., Cruz, T.B.: Automated identification of
  hookahs (waterpipes) on instagram: an application in feature extraction using
  convolutional neural network and support vector machine classification.
  Journal of Medical Internet Research  \textbf{20}(11),  e10513 (2018)

\bibitem{zhang2019modified}
Zhang, Y., Davison, B.D.: Modified distribution alignment for domain adaptation
  with pre-trained inception resnet. arXiv preprint arXiv:1904.02322  (2019)

\bibitem{zhang2020impact}
Zhang, Y., Davison, B.D.: Impact of imagenet model selection on domain
  adaptation. In: Proceedings of the IEEE Winter Conference on Applications of
  Computer Vision Workshops. pp. 173--182 (2020)

\bibitem{zhang2019domain}
Zhang, Y., Tang, H., Jia, K., Tan, M.: Domain-symmetric networks for
  adversarial domain adaptation. In: Proceedings of the IEEE Conference on
  Computer Vision and Pattern Recognition. pp. 5031--5040 (2019)

\bibitem{zhang2019transductive}
Zhang, Y., Xie, S., Davison, B.D.: Transductive learning via improved geodesic
  sampling. In: Proceedings of the 30th British Machine Vision Conference
  (2019)

\bibitem{zhang2020domain}
Zhang, Y., Davison, B.D.: Domain adaptation for object recognition using
  subspace sampling demons. Multimedia Tools and Applications  (2020)

\bibitem{zoph2018learning}
Zoph, B., Vasudevan, V., Shlens, J., Le, Q.V.: Learning transferable
  architectures for scalable image recognition. In: Proceedings of the IEEE
  conference on computer vision and pattern recognition. pp. 8697--8710 (2018)

\end{thebibliography}

\end{document}